\documentclass{article}
\usepackage{fullpage}
\usepackage{graphicx}  	
\usepackage{amssymb}
\usepackage{amsmath}  
\usepackage{amstext}  
\usepackage{amsthm}
\usepackage{natbib}
\usepackage[flushleft]{threeparttable} 
\usepackage{float} 
\usepackage{hyperref} 
\newtheorem{definition}{Definition}
\usepackage{algorithm,algorithmic}
\newtheorem{thm}{Theorem}

\newtheorem{ass}{Assumption}

\hypersetup{colorlinks = true, citecolor = blue}  
\DeclareMathOperator*{\argmin}{arg\,min} 

\usepackage{subfigure}
\def \E {\mathrm{E}}
\def \g {\mathbf{g}}

\def \z {\mathbf{z}}
\def \u {\mathbf{u}}

\def \R {\mathbb{R}}
\def \S {\mathcal{S}}

\def \N {\mathcal{N}}

\def \v {\mathbf{v}}

\def \P {\mathcal{P}}

\def \S {\mathcal{S}}
\def \E {\mathrm{E}}
\def \v {\mathbf{v}}

\def \g {\mathbf{g}}
\def \R {\mathbb{R}}
\def \u {\mathbf{u}}

\def \x {\mathbf{x}}
\def \X {\mathcal{X}}
\def \Y {\mathcal{Y}}

\def \w {\mathbf{w}}

\def \rh {\widehat r}
\def \G {\mathcal{G}}
\def \Z {\mathcal{Z}}

\begin{document}
\title{\bf Why Does Multi-Epoch Training Help?}
\author{Yi Xu, Qi Qian, Hao Li, Rong Jin\\ 
Machine Intelligence Technology, Alibaba Group\\
\{yixu, qi.qian, lihao.lh,  jinrong.jr\}@alibaba-inc.com
}
\date{First Version: \today
}
\maketitle

\begin{abstract}
Stochastic gradient descent (SGD) has become the most attractive optimization method in training large-scale deep neural networks due to its simplicity, low computational cost in each updating step, and good performance. Standard excess risk bounds show that SGD only needs to take one pass over the training data and more passes could not help to improve the performance. Empirically, it has been observed that SGD taking more than one pass over the training data (multi-pass SGD) has much better excess risk bound performance than the SGD only taking one pass over the training data (one-pass SGD). However, it is not very clear that how to explain this phenomenon in theory. In this paper, we provide some theoretical evidences for explaining why multiple passes over the training data can help improve  performance under certain circumstance. Specifically, we consider smooth risk minimization problems whose objective function is non-convex least squared loss. Under Polyak-\L ojasiewicz (PL) condition, we establish faster convergence rate of excess risk bound for multi-pass SGD than that for one-pass SGD. 
\end{abstract}

\section{Introduction}
We consider the following stochastic optimization problem in this work:
\begin{align}\label{opt:prob:empr}
   \min_{\w\in\R^d} F_\S(\w) := \frac{1}{n}\sum_{i=1}^{n}\ell(\w;\z_i),
\end{align}
where $\w$ is the model parameters, $\S=\{\z_1,\dots,\z_n\}$ is a set of training examples sampled from a  distribution $\P$ with support on $\Z$, and $\ell(\w;\z):\R^d \times \Z \longrightarrow \R_+$ is a non-negative smooth loss function. Problem (\ref{opt:prob:empr}) arises in most machine learning based optimization problems such as standard {\bf empirical risk minimization (ERM)} and deep learning. 

Although tremendous stochastic algorithms have been proposed to solving problem (\ref{opt:prob:empr}), stochastic gradient descent (SGD)~\citep{robbins1951stochastic} is probably the simplest and has emerged as one of the most popular techniques. Specifically, at each training iteration $t$,  SGD randomly samples an example $\z_{i_t}$ and then it updates solutions iteratively by
\begin{align}\label{update:sgd}
    \w_{t+1} = \w_t - \eta_t\g_t, 
\end{align}
where $\g_t := \nabla_\w \ell(\w_t;\z_{i_t})$ is the stochastic gradient of objective function at $\w_t$, $i_t$ is uniformly sampled from $[n]$, $\eta_t>0$ is a learning rate. Theoretical convergences of SGD and its variants from an optimization viewpoint have been extensively studied under different conditions of the objection function~\citep{nemirovski2009robust,ghadimi2013stochastic,karimi2016linear,xu2017stochastic}. In real applications such as training deep neural networks, we are more interested in 
whether the learned model can generalize well and result in the least excess risk~\citep{vapnik2013nature,bottou2007tradeoffs}? This introduces us another direction of research, which is the generalization of SGD. That is to say, we want to test the performance of the output of SGD by solving problem (\ref{opt:prob:empr}) on the following {\bf expected risk minimization} (a.k.a. population risk minimization) 
\begin{align}\label{opt:prob}
   \min_{\w\in\R^d} F(\w) := \E_{\z\sim\P}[\ell(\w;\z)].
\end{align}
In this work, we consider the excess risk as the performance measurement
\begin{align}\label{er}
   F(\widehat\w) - F(\w_*),
\end{align}
where $\w_* \in\argmin_{\w\in\R^d} F(\w)$ is the optimal solution, $\widehat\w$ is the output of SGD for solving problem (\ref{opt:prob:empr}). The upper bound of excess risk (\ref{er}) is known as {\bf excess risk bound}, which has been extensively studied in the literature. In the seminal paper by~\cite{hardt2016train}, the authors establish standard excess risk bound $O(1/\sqrt{n})$ under the convex optimization setting, then \cite{zhou2018generalization} extend the result of excess risk bound $O(1/\sqrt{n})$ to non-convex optimization setting. With different assumptions such as strongly convex, Polyak-\L ojasiewicz (PL)~\citep{polyak1963gradient} and the quadratic growth (QG) conditions, many works (e.g., \citep{charles2018stability,kuzborskij2018data,yuan2019stagewise,lei2020fine,leisharper2021}) achieve fast rate of excess risk bound $O(1/n)$. One interesting observation from the theoretical side is that SGD only needs to take $O(n)$ stochastic gradient evaluations to get this fast rate $O(1/n)$, meaning that one-pass SGD can achieve the fast rate under some assumptions on the objective function. However, in practice like training neural networks~\citep{he2019bag}, continuing running SGD after first epoch can still reduce the testing error, which makes multi-pass SGD more powerful than one-pass SGD. Several works have been studied the generalization of multi-pass SGD and fast rates are established under different conditions~\citep{rosasco2015learning,lin2016generalization,lin2017optimal,pillaud2018statistical,feldman2019high,lei2021generalization}. In general, existing fast rates are established in the convex optimization setting and they are not better than the rate of $O(1/n)$. It not fully explains the phenomenon observed in practice for deep learning that continuing training SGD for more epochs can still improve the performances on testing data. In this work, we aim to theoretically understand the observation and try to answer 
the following question:
{
\begin{center}
   \bf Why multiple epochs can help improve the generalization performances? 
\end{center}
}
To answer this question, we will establish a faster excess risk bound for non-convex SGD. When the objective function is smooth and satisfies PL condition, we will show that SGD can achieve the excess risk bound $\widetilde O(1/n^2)$ by continuing conducting SGD after the first epoch. This reveals that multi-pass on training data when using SGD for training deep learning would help to improve the performance on testing data. It is worth mentioning that beyond using SGD, many existing works provide faster rates of excess risk bound under different convex settings~\citep{zhang2017empirical,liu2018fast,ye2019fast}. Their excess risk bounds are constructed for the optimal solution of (\ref{opt:prob:empr}) and they did not consider an efficient optimization algorithm such as SGD for solving their formulation. By contrast, we focus on the excess risk bound for non-convex SGD in this work.

\section{Preliminaries and Notations}
Let the pair of input $\x\in\X$ and label $y\in\Y$ follows a distribution $\P$: $\z = (\x,y) \sim \P$. Let denote by $f(\w;\x): \R^d\times\X \to \Y$ the prediction function parameterized by $\w\in\R^d$, and we want it is as close as possible to label $y$. Based on the given data set  $\S=\{\z_1,\dots,\z_n\}$ that is sampled from the distribution $\P$, we are interested in solving the empirical risk minimization problem (\ref{opt:prob:empr}). Following by~\citep{allen2019convergence}, we consider non-convex squared loss in this work, whose function is given by
\begin{align}\label{loss:LSE}
\ell(\w;\z) := \frac{1}{2}(f(\w;\x)-y)^2.
\end{align} 
Since the non-convexity of $f(\w;\x)$, the loss function $\ell(\w;\z)$ is non-convex. 

Let denote by $\nabla F(\w)$ the gradient of a function $F(\w)$. We use $\|\cdot\|$ to denote the Euclidean norm and use $\langle \cdot,\cdot\rangle$ to denote the inner product. Let $\E_\z[\cdot]$ be the expectation taking over random variable $\z$.

Next, we present some assumptions that will be used in the convergence analysis in this work. Specifically, we make the following assumptions for the loss function.
\begin{ass}\label{ass:1}
Assume the following conditions hold: 
\begin{itemize}
\item[(i)]  The stochastic gradient of $F(\w)$ is unbiased, i.e., $\E_{\z}[\nabla \ell(\w;\z)] = \nabla F(\w)$, and the variance of stochastic gradient is bounded. Specifically, theres exists two constants $B>0$ and $G>0$, such that $$|y|+|f(\w;\x)|\le B,\quad \|\nabla f(\w;\x)\|\le G.$$  
\item[(ii)] $F(\w)$ is smooth with an $L$-Lipchitz continuous gradient, i.e., it is differentiable and there exists a constant $L>0$ such that $$\|\nabla F(\w)  - \nabla F(\u)\|\leq L\|\w - \u\| ,\forall \w, \u \in\R^d.$$
\end{itemize}
\end{ass}
Assumption~\ref{ass:1} (i) and (ii) are commonly used assumptions in the literature~\citep{ghadimi2013stochastic,rosasco2015learning,lin2016generalization,lin2017optimal,pillaud2018statistical,yuan2019stagewise}.
Assumption~\ref{ass:1} (i) assures that the stochastic gradient is bounded, i.e., $\|\nabla\ell(\w;\z)\|\le GB$. Thus, by Jensen's inequality we have
$\E_{\z}\left[\left\|\nabla \ell(\w;\z) - \nabla F(\w)\right\|^2\right] \le \sigma^2$, where $\sigma^2 := 4G^2B^2$. 
Assumption~\ref{ass:1} (ii) says the objective function is $L$-smooth, and it has an equivalent expression~\citep{opac-b1104789} which is $$F(\w) - F(\u) \le \langle \nabla F(\u), \w - \u \rangle + \frac{L}{2}\|\w-\u\|^2, \forall \w, \u \in \R^d.$$   
We then introduce two functional properties as follows.
\begin{definition}[Strongly convex]\label{def:strong:cvx}
We say $F(\w)$ is strongly convex, if
there exists a constant $\gamma>0$ such that $F(\w) - F(\u) \ge \langle \nabla F(\u), \w - \u \rangle + \frac{\gamma}{2}\|\w-\u\|^2, \forall \w, \u \in \R^d.$
\end{definition}
\begin{definition}[Polyak-\L ojasiewicz (PL) condition]\label{def:PL}
There exists a constant $\mu>0$ such that $2\mu(F(\w)- F_*) \le \|\nabla F(\w)\|^2, \forall \w\in\R^d$, where $F_* = \min_{\w\in\R^d} F(\w)$ is the optimal value.
\end{definition}
The PL condition has been theoretically and empirically observed in training deep neural networks~\citep{allen2019convergence, yuan2019stagewise}. This condition is widely used to establish convergence in the literature of non-convex optimization, please see~\citep{yuan2019stagewise, karimi2016linear, li2018simple, charles2018stability} and references therein.

\section{Main Results}\label{sec:main:res}
We will present the main theoretical results in this section. The analyses contain two parts. First, we focus on the optimization of the first epoch. Note that in the first epoch, it satisfies the condition of statistical independence and therefore we can apply the standard analysis of stochastic gradient descent. Second, we consider the updates after the first epoch. In our analysis, we set $
F(\w_*) = 0$ without loss of generality, a common property observed in training deep neural networks~\citep{zhang2016understanding,allen2019convergence,du2018gradient,du2019gradient,arora2019fine,chizat2019lazy, hastie2019surprises,yun2019small, zou2020gradient}. 

First, we will show that under the PL-condition in Definition~\ref{def:PL}, if we use a uniformly sampling without replacement, one-pass SGD has the excess risk bound of $\widetilde O(1/n)$. We present the results in the following theorem.
\begin{thm}[First Epoch]\label{thm:non:convex:first:epoch}
Suppose the objective function $F$ satisfies Assumption~\ref{ass:1} and the PL-condition in Definition~\ref{def:PL}, 
if the learning rate is set to be $\eta = \frac{1}{\mu n} \log\left(\frac{n\mu^2F(\w_0)}{\sigma^2 L}\right)$ in SGD, then we have
\begin{align}\label{eq:thm:non:convex:first:epoch}
  \E[F(\w_n)] \le O\left( \frac{\sigma^2L\log(n)}{\mu^2 n}\right).
\end{align}
\end{thm}

Next, let’s examine the late stage of the optimization beyond the first epoch, in which we do not have the statistical independence between the model to be updated and the training data. As a result, our objective function ``becomes" $F_\S(\w)$. 
For any iteration $t$ after first epoch (we denote by $\w_1$ the solution obtained from the first epoch), define
\begin{align}
    \Omega(\rh_t,s) = \left\{\w: \frac{1}{n}\sum_{i=1}^n\left(f(\w;\x_i) - y_i\right)^2 \leq \rh_t^2, \quad \frac{1}{n}\sum_{i=1}^n\left|f(\w;\x_i) - y_i\right| \leq \sqrt{s_t} \right\}
\end{align}
where
\begin{align}
\rh_t^2 := \frac{1}{n}\sum_{i=1}^n\left(f(\w_t;\x_i) - y_i\right)^2,\quad \sqrt{s_t} := \sqrt{\frac{s}{n}} \rh_t. 
\end{align}
We assume that $s \ll n$, implying that vector $(f(\w_t;\x_1)-y_1, \ldots, f(\w_t;\x_n)-y_n)$ is a sparse vector. This sparse assumption can be relaxed when the objective function has an additional property -- strongly convex. In such setting, strongly convex can be considered as a special case of PL condition (see the Proof of Theorem~\ref{thm:strongly:convex:first:epoch} in Appendix). For completeness, we include the analysis and convergence results for strongly convex case in Appendix.

We give the following results of bounding the variances of the stochastic gradients, which is the key to the analyses.
\begin{thm}\label{thm:non:convex:key:1}
Under the assumption $n \geq 64B^2\log(2/\delta)$, we have, with a probability $1 - 3\delta$, 
\begin{align}\label{eq:thm:5:key:1}
\max\limits_{\w \in \Omega(\rh, s)} \| \nabla F(\w) - \nabla F_\S(\w) \| 
\leq  C\left(\frac{GB\log(1/\delta)}{n} + G\rh\left[\sqrt{\frac{\log(1/\delta)}{n}} + \sqrt{\frac{s}{n}\log\frac{2m}{s}}\left(1 + \log n\right) \right] \right),
\end{align}
where $C$ is a universal constant. 
\end{thm}

Now, we are ready to give the main convergence result for the epochs after the first one. 
\begin{thm}[After First Epoch]\label{thm:non:convex:after:first:epoch}
Suppose the objective function $F$ satisfies Assumption~\ref{ass:1} and the PL-condition in Definition~\ref{def:PL}, 
if $n \ge \frac{6C^2G^2}{\mu} \left[ \log(1/\delta)  + s\log\left(\frac{2n}{s}\right)\left(1 + \log n\right)^2 \right]$, by setting $t = \frac{2}{\eta\mu}\log\left( \frac{\mu F(\w_1)}{\eta G^2B^2L}\right)$ and $\eta = \min\left( \frac{1}{n^2}, \frac{1}{L}\right)$, then with probability at least $1-\delta$  we have
\begin{align}\label{eq:thm:non:convex:after:first:epoch}
F(\w_{t+n})
\leq & \frac{3 C^2G^2B^2\log^2\left(\frac{2}{\eta\mu\delta} \log\left( \frac{\mu F(\w_1)}{\eta G^2B^2 L}\right) \right)}{n^2\mu} + \frac{2\eta G^2B^2 L}{\mu} \le \widetilde O\left( \frac{1}{n^2}\right).
\end{align}
\end{thm}
{\bf Remark.} Based on Theorem~\ref{thm:non:convex:after:first:epoch}, we know the excess risk bound is $\widetilde O(1/n^2)$ if SGD takes $n$ passes over the training data whose sample size is $n$. In particular, the excess risk bound is in the order of $\widetilde O\left(1/n^2 + \eta \right)$ when $t=\widetilde O(1/\eta)$, that is, if SGD continutes running $\widetilde O(1/\eta) \le n^2$ iterations, the excess risk bound is $\widetilde O\left(\eta \right)$. However, after $O(n^2)$ iterations ($O(n)$ epochs), the excess risk bound can not be further improved even SGD runs more epochs.

\section{Proofs}\label{sec:proofs}
In this section, we provide the proofs of theorems in Section~\ref{sec:main:res}.
\subsection{Proof of Theorem~\ref{thm:non:convex:first:epoch}}
We first focus on the optimization of the first epoch. Note that in the first epoch, it satisfies the condition of statistical independence and therefore we can apply the standard analysis of stochastic gradient descent. Then, for each iteration $t$, we have
\begin{align}
  \nonumber &\E[F(\w_{t+1}) - F(\w_t)] \\
  \nonumber\overset{(a)}{\le}& \E[\langle \nabla F(\w_t), \w_{t+1} - \w_t \rangle] + \frac{L}{2}\E[\|\w_{t+1}-\u_t\|^2]\\
  \nonumber\overset{(b)}{\le}& -\eta\left(1-\frac{\eta L}{2}\right) \E[\| \nabla F(\w_t)\|^2] + \frac{\eta^2\sigma^2L}{2}\\
  \overset{(c)}{\le} & - \eta\mu \E[F(\w_t)-F_*] + \frac{\eta^2\sigma^2L}{2},
\end{align}
where (a) uses the smoothness of $F(\w)$ in Assumption~\ref{ass:1} (ii); (b) uses $\w_{t+1} = \w_t - \eta \g_t$, and $\E[\g_t] = \nabla F(\w_t)$ by the unbiased property of $\nabla \ell(\w;\z)$ in Assumption~\ref{ass:1} (i), and $\E[\|F(\w_t) - \g_t\|^2]\le\sigma^2$ in Assumption~\ref{ass:1} (ii); (c) uses $\eta L \le 1$ and the PL-condition in Definition~\ref{def:PL}. Therefore, we have
\begin{align}
  \E[F(\w_{n})-F_*] \le\exp(- \eta\mu n) F(\w_0) + \frac{\eta\sigma^2L}{2\mu} \le O\left( \frac{\sigma^2L\log(n)}{\mu^2 n}\right),
\end{align}
where the last inequality is due to the setting of $\eta = \frac{1}{\mu n} \log\left(\frac{n\mu^2F(\w_0)}{\sigma^2 L}\right)$. We complete the proof by using the fact that $F_*=0$.

\subsection{Proof of Theorem~\ref{thm:non:convex:key:1}}
In this proof, we consider the case of iteration $t+1$ conditional on iteration $t$, then for simplicity, we omit the subscript $t$. Since we have
\[
\max_{\w \in \Omega(\rh,s)} \|\nabla F(\w) - \nabla F_\S(\w)\| = \max_{\w \in \Omega(\rh,s)}\left\|\frac{1}{n}\sum_{i=1}^n (f(\w; \x_i) - y_i)\nabla f(\w; \x_i) - \nabla F(\w)\right\|,
\]
using the Bousquet inequality~\citep{koltchinskii2011oracle}, with appropriate definition of the function space, we have, with a probability $1 - \delta$
\begin{eqnarray*}
\lefteqn{\max\limits_{\w \in \Omega(\rh,s)} \left\|\frac{1}{n G}\sum_{i=1}^n (f(\w; \x_i) - y_i)\nabla f(\w; \x_i) - \nabla F(\w)\right\|} \\
& \leq & 2\E\left[\max\limits_{\w \in \Omega(\rh,s)} \left\|\frac{1}{n G}\sum_{i=1}^n (f(\w; \x_i) - y_i)\nabla_k f(\w; \x_i) - \nabla_k F(\w)\right\|\right] + \frac{4B\log(1/\delta)}{3n} + \sqrt{\frac{2\sigma_P^2}{n}\log\frac{1}{\delta}} \end{eqnarray*}
where
$\sigma_P^2 \leq \frac{1}{G^2}\max\limits_{\w \in \Omega(\rh, s)}\E\left[(f(\w; \x_i) - y_i)^2\|\nabla f(\w; \x_i)\|^2\right] \leq \max\limits_{\w \in \Omega(\rh,s)}\E\left[(f(\w;\x_i) - y)^2\right]\le r^2$ with the Assumption~\ref{ass:1} that $\|\nabla f(\w;\x)\|\le G$.
Using the Berstein inequality, we have, with a probability $1 - \delta$, 
\[
\rh^2 \geq r^2 - \frac{2B^2}{3n}\log\frac{2}{\delta} - 2B\sqrt{\frac{\rh^2}{n}\log\frac{2}{\delta}}
\]
Using the assumption that $m \geq 64B^2\log(2/\delta)$, we have, with a probability $1 - \delta$
\[
r^2 \leq 2\rh^2 + \frac{4B^2}{3n}\log\frac{2}{\delta}, 
\]
and therefore
\[
\sigma_P^2 \leq 2\rh^2 + \frac{4B^2}{3n}\log\frac{2}{\delta}.
\]
As a result, with a probability $1 - 2\delta$, we have
\begin{align*}
&\max\limits_{\w \in \Omega(\rh,s)} \left\|\frac{1}{n G}\sum_{i=1}^n (f(\w; \x_i) - y_i)\nabla f(\w; \x_i) - \nabla F(\w)\right\| \\
 \leq & 2\E\left[\max\limits_{\w \in \Omega(\rh,s)} \left\|\frac{1}{n G}\sum_{i=1}^n (f(\w; \x_i) - y_i)\nabla_k f(\w; \x_i) - \nabla_k F(\w)\right\|\right] + \frac{3B}{n}\log\frac{1}{\delta} + 2\rh \sqrt{\frac{2\log(1/\delta)}{n}} \\
\leq & 4\E_{x,\sigma}\left[\max\limits_{\w \in \Omega(\rh,s)} \left\|\frac{1}{n G}\sum_{i=1}^n \sigma_i (f(\w; \x_i) - y_i) \nabla f(\w; \x_i)\right\|\right] + \frac{4B\log(1/\delta)}{n} + 2\rh\sqrt{\frac{2\log(1/\delta)}{n}}.
\end{align*}
We then bound $\E_{x,\sigma}[\cdot]$. Using Klei-Rio bound~\citep{koltchinskii2011oracle}, with a probability $1 - 2\delta$, we have
\begin{eqnarray*}
\lefteqn{\E_{x,\sigma}\left[\max\limits_{\w \in \Omega(\rh,s)} \left\|\frac{1}{n}\sum_{i=1}^n \sigma_i (f(\w; \x_i) - y_i)\nabla f(\w; \x_i)\right\|\right]} \\
& \leq & 2\E_{\sigma}\left[\max\limits_{\w \in \Omega(\rh,s)} \left\|\frac{1}{n}\sum_{i=1}^n \sigma_i (f(\w; \x_i) - y_i)\nabla f(\w; \x_i) \right\|\right] + 2\rh\sqrt{\frac{\log(1/\delta)}{n}} + \frac{9B}{n}\log\frac{1}{\delta}
\end{eqnarray*}
We can bound the expectation term using Dudley entropy bound (Lemma A.5 of~\citep{bartlett2017spectrally}),i.e.
\begin{align}
\nonumber&\E_{\sigma}\left[\max\limits_{\w \in \Omega(\rh,s)} \left\|\frac{1}{n GB}\sum_{i=1}^n \sigma_i (f(\w; \x_i) - y_i)\nabla f(\w; \x_i) \right\|\right] \\
\nonumber \leq & \E_{\sigma}\left[\max\limits_{\w \in \Omega(\rh,s), h \in S_d} \left\|\frac{1}{n GB}\sum_{i=1}^n \sigma_i (f(\w; \x_i) - y_i)\langle \nabla f(\w; \x_i), h\rangle \right\|\right] \\
\leq & \inf\limits_{\alpha} \left(\frac{4\alpha}{\sqrt{n}} + \frac{12}{n}\int_\alpha^{\sqrt{n}}\sqrt{\log\N(\G(\rh,s);\epsilon)} d\epsilon \right)
\end{align}
where $\N(\G(\rh,s); \epsilon)$ is the covering number of proper $\epsilon$-net over $\G(\rh,s)$ and $S_d$ is a simplex set. Here $\G(\rh,s)$ is defined as
\[
\G(\rh,s) = \left\{v \in \R^n: \|v\| \leq \frac{\sqrt{n}\rh}{B}, \|v\|_1 \leq \frac{\sqrt{n s} \rh}{B} \right\}
\]
Using the covering number result for sparse vectors (Lemma 3.4 of~\citep{plan2013one}), we have
\[
    \log \N\left(\G(\rh,s);\frac{\sqrt{n}\rh}{B}\epsilon\right) \leq \frac{Cs}{\epsilon^2}\log\frac{2n}{s}
\]
and therefore
\[
    \log \N\left(\G(\rh,s);\epsilon\right) \leq \frac{Cn\rh^2s}{B^2\epsilon^2}\log\frac{2n}{s}
\]
Plugging into the Dudley entropy bound, we have
\[
\E_{\sigma}\left[\max\limits_{\w \in \Omega(\rh,s)} \left\|\frac{1}{n B}\sum_{i=1}^n \sigma_i (f(\w; \x_i) - y_i) \right\|\right] \leq \inf\limits_{\alpha}\left(\frac{4\alpha}{\sqrt{n}} + \frac{12\rh}{B}\sqrt{\frac{Cs}{n}\log\frac{2n}{s}}\log\frac{\sqrt{n}}{\alpha}\right)
\]
By choosing $\alpha$ as
\[
    \alpha = \frac{3\rh}{B}\sqrt{Cs\log\frac{2n}{s}},
\]
we have
\[
\E_{\sigma}\left[\max\limits_{\w \in \Omega(\rh,s)} \left\|\frac{1}{n}\sum_{i=1}^n\sigma_i (f(\w; \x_i) - y_i) \right\|\right] \leq 12\rh\sqrt{\frac{Cs}{n}\log\frac{2n}{s}}\left(1 + \frac{1}{2}\log\left(\frac{n}{9\sqrt{Cs\log(2n/s)}}\right)\right)
\]
Combining the above results together, with a probability $1 - 3\delta$, we prove the inequality (\ref{eq:thm:5:key:1}).

\subsection{Proof of Theorem~\ref{thm:non:convex:after:first:epoch}}
In this proof, we consider the updates after first epoch, i.e., $t\ge n$.
\begin{align}\label{thm:non:convex:after:first:epoch:eq:1}
  \nonumber &F(\w_{t+1}) - F(\w_t) \\
  \nonumber\overset{(a)}{\le}& \langle \nabla F(\w_t), \w_{t+1} - \w_t \rangle + \frac{L}{2}\|\w_{t+1}-\w_t\|^2\\
  \nonumber\overset{(b)}{=}& -\eta\langle \nabla F(\w_t), \nabla \ell(\w_t;\z_{i_t}) \rangle + \frac{\eta^2L}{2}\|\nabla \ell(\w_t;\z_{i_t})\|^2\\
  \overset{(c)}{\le} &-\eta \langle \nabla F(\w_t), \nabla \ell(\w_t;\z_{i_t}) \rangle + \frac{\eta^2LG^2B^2}{2},
\end{align}
where (a) uses the smoothness of $F(\w)$ in Assumption~\ref{ass:1} (ii); (b) uses $\w_{t+1} = \w_t - \eta \g_t$; (c) uses Assumption~\ref{ass:1} (i) that $\|\g_t\| \le GB$.
By summing the above inequalities across $i_t = 1,\dots, n$ and dividing by $n$, we get
\begin{align}\label{thm:non:convex:after:first:epoch:eq:2}
  \nonumber &F(\w_{t+1}) - F(\w_t) \\
 \nonumber  \le & -\eta\langle \nabla F(\w_t), \nabla F_\S(\w_t)) \rangle + \frac{\eta^2LG^2B^2}{2},\\
\nonumber \leq & \frac{\eta}{2}\left\|\nabla F(\w_t) - \nabla F_{\S}(\w_t)\right\|^2  - \frac{\eta}{2}\left\|\nabla F(\w_t) \right\|^2 + \frac{\eta^2LG^2B^2}{2}\\
\leq & \frac{\eta M_t}{2} - \eta\mu F(\w_t) + \frac{\eta^2LG^2B^2}{2},
\end{align}
where the last inequality uses PL-condition and Theorem~\ref{thm:non:convex:key:1} with
\begin{align*}
    M_t:=& C^2\left(\frac{GB\log(1/\delta)}{n} + G\rh_t\left[\sqrt{\frac{\log(1/\delta)}{n}} + \sqrt{\frac{s}{n}\log\left(\frac{2n}{s}\right)}\left(1 + \log n\right) \right] \right)^2.
\end{align*}
By assuming that $n$ is large such that $n \ge \frac{6C^2G^2}{\mu} \left[ \log(1/\delta) + s\log\left(\frac{2n}{s}\right)\left(1 + \log n\right)^2 \right]$, then
\begin{align*}
    M_t \le & \frac{3C^2G^2B^2\log^2(1/\delta)}{n^2} + 3C^2G^2\rh_t^2 \left[ \frac{\log(1/\delta)}{n} + \frac{s}{n}\log\left(\frac{2n}{s}\right)\left(1 + \log n\right)^2 \right] \\
    \le & \frac{3C^2G^2B^2\log^2(1/\delta)}{n^2} + \frac{\mu}{2} \rh_t^2.
\end{align*}
Therefore, with probability at least $1-3\delta$ we have
\begin{align*}
F(\w_{t+1}) - F(\w_t)
\leq & \frac{3\eta C^2G^2B^2\log^2(1/\delta)}{2n^2} + \frac{\eta\mu}{2}  \frac{1}{2} \rh_t^2- \eta\mu F(\w_t) + \frac{\eta^2 G^2B^2 L}{2}\\
\leq & \frac{3\eta C^2G^2B^2\log^2(1/\delta)}{2n^2} - \frac{\eta\mu}{2}F(\w_t) + \frac{\eta^2 G^2B^2 L}{2},
\end{align*}
which implies that with probability at least $1-3t\delta$,
\begin{align*}
F(\w_{t+1})
\leq & \left(1- \frac{\eta\mu}{2}\right)^tF(\w_1) + \frac{3 C^2G^2B^2\log^2(1/\delta)}{n^2\mu} + \frac{\eta G^2B^2 L}{\mu}\\
\leq &\exp\left(- \frac{\eta\mu t}{2}\right)F(\w_1) + \frac{3 C^2G^2B^2\log^2(1/\delta)}{n^2\mu} + \frac{\eta G^2B^2 L}{\mu}.
\end{align*}
By setting $t = \frac{2}{\eta\mu}\log\left( \frac{\mu F(\w_1)}{\eta G^2B^2 L}\right)$ and $\eta = \min\left( \frac{1}{n^2}, \frac{1}{L}\right)$, then with probability at least $1-\delta$  we have
\begin{align*}
F(\w_{t+1})
\leq & \frac{3 C^2G^2B^2\log^2\left(\frac{2}{\eta\mu\delta} \log\left( \frac{\mu F(\w_1)}{\eta G^2B^2 L}\right) \right)}{n^2\mu} + \frac{2\eta G^2B^2 L}{\mu} \le \widetilde O\left( \frac{1}{n^2}\right).
\end{align*}

\section{Conclusions}
In this work, we analyze the convergence of multi-pass SGD for solving non-convex smooth risk minimization problem. Under the Polyak-\L ojasiewicz condition, we establish faster rate of excess risk bound for multi-pass SGD than one-pass SGD. This result reveals that multiple epochs in training deep neural networks can help improve the generalization performance. However, multi-pass SGD can not always improve the performance after it takes $n$ passed over the training data, where $n$ is the sample size of training data.

\bibliographystyle{plainnat} 
\bibliography{ref}

\newpage
\appendix
\section{Results for Strongly Convex}
In this section, we will show the results under the strongly convex assumption in Definition~\ref{def:strong:cvx}. First, i                  f we use a uniformly sampling without replacement, SGD can have the generalization bound of $\widetilde O(1/n)$ by passing training data only once. We present the results in the following theorem.
\begin{thm}[First Epoch]\label{thm:strongly:convex:first:epoch}
Suppose the objective function $F$ satisfies Assumption~\ref{ass:1} and the strongly convex assumption in Definition~\ref{def:strong:cvx}, if the learning rate is set to be $\eta = \frac{4L}{\gamma^2 n} \log\left(\frac{n\gamma^4F(\w_0)}{4\sigma^2 L^2}\right)$ in SGD, then we have
\begin{align}\label{eq:thm:strongly:convex:first:epoch}
  \E[F(\w_n)] \le O\left( \frac{\sigma^2L^3\log(n)}{\gamma^4 n}\right).
\end{align}
\end{thm}

Next, let’s examine the late stage of the optimization beyond the first epoch, in which we do not have the statistical independence between the model to be updated and the training data. As a result, our objective function ``becomes" $F_\S(\w)$. 
For any iteration $t$ after first epoch (we denote by $\w_1$ the solution obtained from the first epoch), define
\begin{align}
    \Omega(r_t) = \left\{\w: \|\w-\w_*\|\le r_t\right\}
\end{align}
where
\begin{align}
r_t := \left\|\w_t - \w_*\right\|.
\end{align}
We assume that our model is powerful in fitting the data and therefore $\nu$ will be a small value. Specifically, we assume $\nu := \max_{(\x,y)}\|\nabla f(\w_*;\x)(f(\w_*;\x) - y)\|\le Gr$. 

We give the following results of bounding the variances of the stochastic gradients, which is the key to the analyses.
\begin{thm}\label{thm:strongly:convex:key:1}
Under the assumption $\nu := \max_{(\x,y)}\|\nabla f(\w_*;\x)(f(\w_*;\x) - y)\|\le Gr$ and $\nu < 1$, we have, with a probability $1 - 3\delta$, 
\begin{align}\label{eq:thm:2:key:1}
\max\limits_{\w \in \Omega(r)} \| \nabla F(\w) - \nabla F_\S(\w) \| 
\leq  C\left(\frac{Gr}{n} + Gr\sqrt{\frac{\log(1/\delta)+d\log(n)}{n}} \right),
\end{align}
where $C$ is a universal constant. 
\end{thm}
Now, we are ready to give the convergence result for the epochs after the first one. 
\begin{thm}[After First Epoch]\label{thm:strongly:convex:after:first:epoch}
Suppose the objective function $F$ satisfies Assumption~\ref{ass:1} and the $\gamma$-strongly convex in Definition~\ref{def:strong:cvx}, if $n \ge \frac{4 C^2G^2(\log(t/\delta)+d\log(n))}{\gamma^2}$,
by setting $t = \frac{2}{\eta\gamma}\log\left( \frac{\gamma n^2F(\w_1)}{2C^2G^2B^2}\right)$ and $\eta = \min\left( \frac{1}{n^2}, \frac{1}{L}\right)$, then with probability at least $1-\delta$  we have
\begin{align}\label{eq:thm:strongly:convex:after:first:epoch}
F(\w_{t+1})\le \frac{4 C^2G^2B^2}{\gamma n^2}+ \frac{\eta G^2B^2 L}{\gamma}\le O\left( \frac{1}{n^2}\right).
\end{align}
\end{thm}

\subsection{Proof of Theorem~\ref{thm:strongly:convex:first:epoch}}
We first focus on the optimization of the first epoch. Note that in the first epoch, it satisfies the condition of statistical independence and therefore we can apply the standard analysis of stochastic gradient descent. Then, for each iteration $t$, we have
\begin{align}\label{thm:1:eqn:1}
   \nonumber &\E[F(\w_{t+1}) - F(\w_t)] \\
   \nonumber\overset{(a)}{\le}& \E[\langle \nabla F(\w_t), \w_{t+1} - \w_t \rangle] + \frac{L}{2}\E[\|\w_{t+1}-\w_t\|^2]\\
   \nonumber\overset{(b)}{\le}& -\eta\left(1-\frac{\eta L}{2}\right) \E[\| \nabla F(\w_t)\|^2] + \frac{\eta^2\sigma^2L}{2}\\
   \overset{(c)}{\le} & - \eta\widehat\gamma \E[F(\w_t)-F_*] + \frac{\eta^2\sigma^2L}{2},
\end{align}
where (a) uses the smoothness of $F(\w)$ in Assumption~\ref{ass:1} (ii); (b) uses $\w_{t+1} = \w_t - \eta \g_t$, and $\E[\g_t] = \nabla F(\w_t)$ by the unbiased property of $\nabla \ell(\w;\z)$ in Assumption~\ref{ass:1} (i), and $\E[\|F(\w_t) - \g_t\|^2]\le\sigma^2$ in Assumption~\ref{ass:1} (ii); (c) uses $\eta L \le 1$ and the $\gamma$-strongly convex in Definition~\ref{def:strong:cvx}, and (\ref{thm:1:eq:3}) with $\widehat\gamma := \frac{\gamma^2}{4L}$. In fact, by the smoothness of $F$, we get $F(\w) - F(\v) \le \langle \nabla F(\v), \w - \v \rangle + \frac{L}{2}\|\w-\v\|^2$. Let $\v = \w_*$, then
\begin{align}\label{thm:1:eq:2}
    F(\w) - F_* \le \frac{L}{2}\|\w-\w_*\|^2.
\end{align}
On the other hand, by the strong convexity of $F$, we get $F(\v) - F(\w) \ge \langle \nabla F(\w), \v - \w \rangle + \frac{\gamma}{2}\|\w-\v\|^2$. Let $\v = \w_*$, then
\begin{align*}
    \|\nabla F(\w)\| \|\w - \w_*\| \ge & \langle \nabla F(\w), \w - \w_* \rangle \\
    \ge & F(\w) - F_* + \frac{\gamma}{2}\|\w-\w_*\|^2 \\
    \ge & \frac{\gamma}{2}\|\w-\w_*\|^2,
\end{align*}
which implies
\begin{align}\label{thm:1:eq:3}
    \|\nabla F(\w)\|^2 \ge \frac{\gamma^2}{4}\|\w-\w_*\|^2 \overset{(\ref{thm:1:eq:2})}{\ge} \frac{\gamma^2}{4} \frac{2(F(\w)-F_*)}{L} = \frac{\gamma^2}{2L}(F(\w)-F_*).
\end{align}
Therefore, by (\ref{thm:1:eqn:1}) we have
\begin{align}\label{thm:1:eq:4}
  \E[F(\w_{n})-F_*] \le\exp(- \eta\widehat\gamma n) (F(\w_0)-F_*) + \frac{\eta\sigma^2L}{2\widehat\gamma} \le O\left( \frac{\sigma^2L\log(n)}{\widehat\gamma^2 n}\right),
\end{align}
where the last inequality is due to the setting of $\eta = \frac{1}{\widehat\gamma n} \log\left(\frac{n\widehat\gamma^2F(\w_0)}{\sigma^2 L}\right)$. We complete the proof by using $F_*=0$ in (\ref{thm:1:eq:4}).

\subsection{Proof of Theorem~\ref{thm:strongly:convex:key:1}}
In this proof, we consider the case of iteration $t+1$ conditional on iteration $t$, then for simplicity, we omit the subscript $t$. To bound $\|\nabla F(\w) - \nabla F_{\S}(\w)\|$, we first have 
\begin{align}
    \|\nabla F(\w) - \nabla F_{\S}(\w)\| \le \|\nabla F(\w) - \nabla F(\w_*) + \nabla F_{\S}(\w_*) - \nabla F_{\S}(\w)\| + \|\nabla F(\w_*) - \nabla F_{\S}(\w_*)\|.
\end{align}
Using the standard concentration inequality, with a probability $1-\delta$, we have
\begin{align}
    \|\nabla F(\w_*) - \nabla F_{\S}(\w_*)\| \le C\left(\frac{\nu}{n} + \nu\sqrt{\frac{\log(1/\delta)}{n}}\right),
\end{align}
where $\nu = \max_{(\x,y)}\|\nabla f(\w_*;\x)(f(\w_*;\x) - y)\|$. 

To bound $\| \nabla F(\w) - \nabla F(\w_*) + \nabla F_{\S}(\w_*) - \nabla F_{\S}(\w)\|$, for any $\w\in\Omega(r)$, we denote by $\N(\Omega(r),\epsilon)$ the covering number of the proper $\epsilon$-net of $\Omega(r)$. {It is easy to verify that with a probability at least $1-\delta$,}
\begin{align}
    \| \nabla F(\w) - \nabla F(\w_*) + \nabla F_{\S}(\w_*) - \nabla F_{\S}(\w)\| \le C\left(\frac{Gr}{n} + Gr\sqrt{\frac{\log(1/\delta)+d\log(r/\epsilon)}{n}}+G\epsilon\right),
\end{align}
where we used the fact that $\N(\Omega(r),\epsilon) = O((r/\epsilon)^d)$. Combining above three inequalities, we have, with a probability at least $1-2\delta$
\begin{align}
    \| \nabla F(\w) - \nabla F_{\S}(\w)\| \le C\left(\frac{\nu}{n}+ \nu\sqrt{\frac{\log(1/\delta)}{n}}+\frac{Gr}{n} + \frac{Gr}{\sqrt{n}}\left(\sqrt{\log(1/\delta)+d\log(n)/2}+1\right)\right),
\end{align}
where we choose $\epsilon=r/\sqrt{n}$.
Since $\nu\le Gr$,  we have, with a probability at least $1-3\delta$
\begin{align}
    \| \nabla F(\w) - \nabla F_{\S}(\w)\|  \le O\left(Gr\sqrt{\frac{\log(1/\delta)+d\log(n)}{n}}\right).
\end{align}

\subsection{Proof of Theorem~\ref{thm:strongly:convex:after:first:epoch}}
In this proof, we consider the updates after first epoch, i.e., $t\ge n$. Following the similar analysis of Theorem~\ref{thm:non:convex:after:first:epoch}, we have
\begin{align}\label{thm:3:eqn:1}
\nonumber F(\w_{t+1}) - F(\w_t) \leq & \frac{\eta}{2}\left\|\nabla F(\w_t) - \nabla F_\S(\w_t)\right\|^2  - \frac{\eta}{2}\left\|\nabla F(\w_t) \right\|^2 + \frac{\eta^2 G^2B^2 L}{2}\\
\leq & \frac{\eta M_t}{2} - \eta\gamma F(\w_t) + \frac{\eta^2 G^2B^2 L}{2},
\end{align}
where the last inequality uses $\gamma$-strongly convexity and Theorem~\ref{thm:strongly:convex:key:1} with
\begin{align}\label{thm:3:eqn:2}
    M_t:=& C^2\left(\frac{Gr_t}{n} + Gr_t\sqrt{\frac{\log(1/\delta)+d\log(n)}{n}}\right)^2.
\end{align}
Therefore, with probability at least $1-3\delta$ we have
\begin{align*}
F(\w_{t+1}) - F(\w_t)
\leq &\eta C^2\left(\frac{G^2B^2}{n^2} + G^2 r_t^2 \frac{\log(1/\delta)+d\log(n)}{n}\right)- \eta\gamma F(\w_t) + \frac{\eta^2 G^2B^2 L}{2}.
\end{align*}
By the strong convexity and $F_*=0$ we know
\begin{align*}
r_t^2 \le \frac{2(F(\w_t))}{\gamma}.
\end{align*}
Thus 
\begin{align*}
 F(\w_{t+1}) - F(\w_t) 
\leq &\frac{\eta C^2G^2B^2}{n^2} +  F(\w_t) \frac{2\eta C^2G^2(\log(1/\delta)+d\log(n))}{\gamma n}- \eta\gamma F(\w_t) + \frac{\eta^2 G^2B^2 L}{2}\\
\leq &\frac{\eta C^2G^2B^2}{n^2} - \frac{\eta \gamma}{2} F(\w_t) + \frac{\eta^2 G^2B^2 L}{2}.
\end{align*}
where the last inequality is due to $n \ge \frac{4 C^2G^2(\log(1/\delta)+d\log(n))}{\gamma^2}$.
Then, with probability at least $1-3t\delta$,
\begin{align*}
F(\w_{t+1})
\leq & \left(1- \frac{\eta\gamma}{2}\right)^tF(\w_1) + \frac{2 C^2G^2B^2}{\gamma n^2}+ \frac{\eta G^2B^2 L}{\gamma}\\
\leq &\exp\left(- \frac{\eta\gamma t}{2}\right)F(\w_1) + \frac{2 C^2G^2B^2}{\gamma n^2}+ \frac{\eta G^2B^2 L}{\gamma}.
\end{align*}
By setting $t = \frac{2}{\eta\gamma}\log\left( \frac{\gamma n^2F(\w_1)}{2C^2G^2B^2}\right)$ and $\eta = \min\left( \frac{1}{n^2}, \frac{1}{L}\right)$, then with probability at least $1-\delta$  we have
\begin{align*}
F(\w_{t+1})
\leq & \frac{4 C^2G^2B^2}{\gamma n^2}+ \frac{\eta G^2B^2 L}{\gamma}\le O\left( \frac{1}{n^2}\right).
\end{align*}

\end{document}